\documentclass[letterpaper, 10 pt, conference]{ieeeconf}  

\IEEEoverridecommandlockouts                              

\usepackage[english]{babel}
\usepackage[utf8]{inputenc}

\usepackage{graphicx,subfigure}
\usepackage{yaacro}

\usepackage{amsmath,amsfonts,amssymb}

\usepackage{amsthm}

\newtheorem{problem}{Problem}

\usepackage{algorithm}
\usepackage{algorithmic}
\usepackage[hidelinks]{hyperref}
\usepackage{comment}
\usepackage{multirow}
\newcommand{\conjunto}[1]{\ensuremath{\mathcal{\uppercase{#1}}}}
\newcommand{\escalar}[1]{\ensuremath{\mathit{#1}}}

\newcommand{\Grafo}{\conjunto{G}}
\newcommand{\ConjVertices}{\conjunto{V}}
\newcommand{\ConjArestas}{\conjunto{E}}
\newcommand{\ConjClusters}{\conjunto{C}}
\newcommand{\ConjSubgrupos}{\conjunto{S}}
\newcommand{\Caminho}{\conjunto{P}}

\newcommand{\vertice}{\escalar{v}}
\newcommand{\aresta}{\escalar{e}}
\newcommand{\cluster}{\escalar{C}}
\newcommand{\subgrupo}{\escalar{S}}
\newcommand{\reward}{\escalar{r}}
\newcommand{\custoAresta}{\escalar{t}}
\newcommand{\noCaminho}{\escalar{p}}

\newcommand{\rewardEarned}{\escalar{R}}
\newcommand{\length}{\escalar{L}}
\newcommand{\runtime}{\escalar{T}}

\newcommand{\visitedCluster}{\escalar{w}}

\newcommand{\budget}{\ensuremath{T_\textrm{max}}}

\usepackage[colorinlistoftodos,prependcaption]{todonotes}

\newcommand{\douglas}[2][orange]{\todo[author=\textbf{Douglas}, inline, color=#1!50]{#2}}

\newcommand{\eg}{\textit{e.g.}}
\newcommand{\ie}{\textit{i.e.}}

\begin{acgroupdef}[list=acronimos]
    \acdef{TSP}{Traveling Salesman Problem}
    \acdef{KP}{Knapsack Problem}
    \acdef{OP}{Orienteering Problem}
    \acdef{COP}{Clustered Orienteering Problem} 
    \acdef{SOP}{Set Orienteering Problem}    
    \acdef{COPS}{Clustered Orienteering Problem with Subgroups}  
    \acdef{PoIs}{Points of Interest}

    \acdef{DOP}{Dubins Orienteering Problem}
    \acdef{OPN}{Orienteering Problem with Neighborhoods}
    \acdef{MEDOP}{Minimal Exposure Dubins Orienteering Problem}
\end{acgroupdef}

\title{\LARGE \bf
Clustered Orienteering Problem with Subgroups
}

\author{Luciano E. Almeida \qquad Douglas G. Macharet
\thanks{This work was supported by CAPES/Brazil - Finance Code 001, CNPq/Brazil, and FAPEMIG/Brazil.}
\thanks{The authors are with the Computer Vision and Robotics Laboratory (VeRLab), Department of Computer Science, Universidade \mbox{Federal} de Minas Gerais, Brazil. E-mails: {\tt\small \{luciano.almeida, doug\}@dcc.ufmg.br}.}
}

\begin{document}

\maketitle
\thispagestyle{empty}
\pagestyle{empty}


\begin{abstract}

This paper introduces an extension to the \ac{OP}, called Clustered Orienteering Problem with Subgroups (COPS). In this variant, nodes are arranged into subgroups, and the subgroups are organized into clusters. A reward is associated with each subgroup and is gained only if all of its nodes are visited; however, at most one subgroup can be visited per cluster. The objective is to maximize the total collected reward while attaining a travel budget.
We show that our new formulation has the ability to model and solve two previous well-known variants, the \ac{COP} and the \ac{SOP}, in addition to other scenarios introduced here.
An Integer Linear Programming (ILP) formulation and a Tabu Search-based heuristic are proposed to solve the problem.
Experimental results indicate that the ILP method can yield optimal solutions at the cost of time, whereas the metaheuristic produces comparable solutions within a more reasonable computational cost.

\end{abstract}


\section{Introduction}

Vehicle routing problems are essential in different research fields, such as Operations Research and Robotics. The \ac{TSP}, in particular, is a well-known combinatorial optimization problem that aims to find the shortest route that visits a given set of cities exactly once and returns to the starting city. Despite being an NP-hard problem, the \ac{TSP} has numerous real-world applications, including logistics, transportation, and VLSI design.

An important generalization of the \ac{TSP} is the \ace{OP}~\cite{golden1987orienteering}, which can be defined as a combination of the \ac{TSP} and a \ac{KP}. In the \ac{OP}, the goal is to determine a route that maximizes a total reward obtained when visiting the cities, with the constraint that it should not exceed a given travel budget (\eg, length or time).

The OP also has several variants, and we are interested in two relatively recent ones, the \ace{COP} and the \ace{SOP}. The \ac{COP}~\cite{Angelelli2014Clustered} considers that the cities are divided into clusters. A reward is associated with each cluster, not the cities individually as in the \ac{OP}, and is collected only if \emph{all cities} belonging to the cluster are visited. The \ac{SOP}~\cite{Archetti2018Set} also considers that the cities are grouped into clusters; however, the reward is collected only if \emph{at least one city} is visited.

\begin{figure}[htpb]
    \centering
    \includegraphics[width=.85\linewidth]{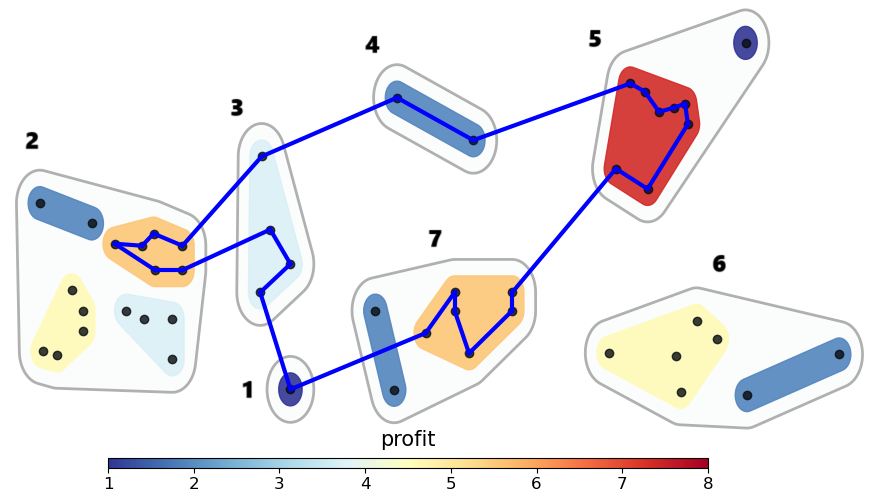}
    \caption{Example of seven clusters with distinct subgroups. The subgroups' associated reward is represented by their colors. The cluster with a single vertex is the start/end point, and the blue line is a possible route.}
    \label{fig:exemplo_ilustrativo}
\end{figure}

In this paper, we propose a novel generalization related to both of these variants, called \ac{COPS}. In the \ac{COPS}, the target locations are also organized into clusters, however, inside a cluster, they are arranged into one or more subgroups, and a reward is assigned to each one of these subgroups. The objective is to maximize the total reward, where a reward is collected if \emph{at most one} subgroup is selected and \emph{all} of its internal locations are visited.
Fig.~\ref{fig:exemplo_ilustrativo} illustrates the problem and shows a possible solution (blue path).


Therefore, in contrast to both \ac{COP} and \ac{SOP}, a cluster can now provide distinct rewards accordingly to the visited subgroup. This formulation is useful in various scenarios, where one can use the clusters to model tasks of a more general mission, while the subgroups define specific ways of executing a task. For example, given a 3D reconstruction problem, a cluster can group all points representing different viewing angles of an object and a subgroup a set of views with certain properties. In this context, a cluster refers to different possible ways of reconstructing the same object.


In summary, our main contributions are:
\begin{itemize}
    \item Concept of a novel variant that allows to model different subgroups in clustered-related OP instances;
    \item Unified formulation for solving both \ac{COP} and \ac{SOP};
    \item Proposal of an exact ILP method;
    \item Design of an efficient Tabu Search-based heuristic.
\end{itemize}




\section{Related Work}



The \acd{OP} is a variant of the \acd{TSP} that focuses on identifying the most rewarding subset of locations to visit with limited resources. This optimization problem has garnered significant attention due to its relevance in various fields, such as logistics, transportation, and tourism.

The classic OP formulation has evolved to encompass a range of generalizations, each tailored to specific characteristics and applications. For example, vehicles with motion constraints, as in the \ac{DOP}~\cite{Penicka2017Dubins}, or the \ac{MEDOP}~\cite{Macharet2021Minimal}, a multi-objective version that handles both curvature and exposure constraints. Moreover, it is worth noting the diversity of variants proposed to tackle specific challenges, such as correlated and time-varying profits~\cite{Yu2016COP, Ma2017SpatioTemporal}, fault-tolerance~\cite{Santos2021Anytime}, environmental properties such as ocean currents~\cite{Mansfield2022Energy}, and region-based demands~\cite{faigl2016self}.

The notion of a \emph{cluster} has been adopted in numerous variations of vehicle routing problems to refer to a subset of customers or locations. Nevertheless, the interpretation of this term and how these target nodes are handled can vary depending on the specific problem under consideration.

In this context, we highlight the \ace{COP}~\cite{Angelelli2014Clustered}, which is the main inspiration for this work. In the \ac{COP}, target locations are grouped into clusters, and all locations in a certain cluster must be visited to collect its associated reward. The original article solves the problem through a branch-and-cut method and Tabu Search algorithms. The COP can be applied to several real-world applications, one such application is in the context of logistics. Consider, for example, the case where customers are clustered into different supply chains, and a company has to design a delivery route that covers most supply chains. Therefore, it might be interesting to design a route that attends all customers in the same area or with similar demands (cluster) before moving to the next region.

The \ac{COP} has also been extended to the case where multiple agents collaborate in the execution of the task of visiting customers~\cite{Yahiaoui2019Clustered}, called Clustered Team Orienteering Problem (CluTOP). In the Clustered Coverage Orienteering Problem (CCOP)~\cite{zhang2020clustered}, the nodes are grouped into clusters representing sampling sites, and a minimum number of nodes must be visited in each cluster so that a reasonable amount of information can be collected about a region.

Another related generalization is the \ace{SOP}, in which the customers are also grouped into clusters, but the reward associated with each cluster is collected by visiting at least one of the customers in the respective cluster. In~\cite{Archetti2018Set}, the authors proposed two strategies to solve the problem, mathematical formulation and a matheuristic algorithm. In~\cite{Penicka2019Variable}, the problem was tackled by both the Variable Neighborhood Search (VNS) method and a novel Integer Linear Programming (ILP) formulation that improves the computational times of discovering the SOP result. Finally, \cite{Gunawan2021Set} proposed a generalization of the SOP where a team of agents performs a set of routes while satisfying travel time limitations and time window constraints. An interesting application for SOP is the same example of logistics and transportation mentioned for COP. However, in this case, instead of having to visit every customer in the chain, the route needs to visit at least one customer, and implicitly, it would be serving the entire chain.

However, in both \ac{COP} and \ac{SOP}, a cluster is commonly regarded as a homogeneous entity that neglects any potential heterogeneity among its constituent nodes. To address this limitation, we propose a novel variant, which designates nodes into two categories: clusters and subgroups. Specifically, a cluster is a collection of one or more subgroups, and a subgroup denotes a set of one or more individual nodes. By introducing this novel formulation, we are able to accommodate various scenarios that were previously infeasible with the COP and SOP, while simultaneously integrating the two problems into a unified framework.

\section{Problem formulation}
\label{section_problem_description}


Given a complete undirected graph $\Grafo = (\ConjVertices, \ConjArestas)$, where
$\ConjVertices$ is the set of vertices $\ConjVertices = \{\vertice_{1}, \vertice_{2},...,\vertice_{n}\}$ and
$\ConjArestas$ the set of edges $\ConjArestas = \{\aresta_{1}, \aresta_{2},...,\aresta_{m}\}$.
%
The vertices are partitioned into subgroups $\ConjSubgrupos = \{\subgrupo_{1}, \subgrupo_{2},..., \subgrupo_{k}\}$, where each vertex is assigned to at least one subgroup $\subgrupo_{i}$. Moreover, it is possible for the same vertex to belong to multiple subgroups.
%
%
Each subgroup $\subgrupo_{i}$ has an associated reward $\reward_{i}$, which is collected only if all vertices of the subgroup are visited.
Each edge $\aresta \in \ConjArestas$ is assigned a traversal cost $\custoAresta_{e}$, which can be determined based on the distance or the time required to traverse it, among other factors. The maximum cost of the vehicle route is limited to $\budget$, \ie, a time or energy budget.



Next, we can define a cluster $\cluster_{g}$ like a set of subgroups $\cluster_{g} \in \ConjClusters = \{\cluster_{1}, \cluster_{2},...,\cluster_{l}\}$. Each subgroup $\subgrupo_{i}$ must belong to at least one cluster $\cluster_{g}$ and the same subgroup can belong to more than one cluster.
%
%
A cluster is considered visited when \emph{exactly} one subgroup belonging to the cluster has been entirely visited.
It is not necessary to visit all vertices of a subgroup in sequence (see cluster \#3 in Fig.~\ref{fig:exemplo_ilustrativo}).


A route starts at $\cluster_{1}$, which must have only one subgroup $\subgrupo_{1}$ with one vertex $\vertice_{1}$, which is the initial vertex, and $\cluster_{end}$ is the cluster that finishes the tour. In this sense, if $\cluster_{1} = \cluster_{end}$ the solution will be a circular route, otherwise the solution will be a non-circular route.



A client $\vertice_{client}$ is a vertex that does not belong to the end cluster or the start cluster. Similarly, we can define $\ConjVertices_{client}$ as the set of client vertices, and $\ConjSubgrupos_{client}$ as the set of subgroups formed only by client vertices.


\begin{problem}[\acd{COPS}]
Given a collection of clusters $\ConjClusters$, each cluster with one or more internal subgroups $\ConjSubgrupos' \subset \ConjSubgrupos$. A subgroup $\subgrupo_{i}$ has an associated reward $\reward_{i}$. The objective is to determine a route that maximizes the total collected reward, which is obtained by visiting all vertices of a subgroup, while respecting a predefined travel budget $\budget$ and the cluster constraint. 
\end{problem}


A valid solution is a route $\Caminho$ composed of a set of vertices $\Caminho = \{\noCaminho_{1},\noCaminho_{2},...,\noCaminho_{w}\}$ that initiates at the start vertex $\noCaminho_{0} = \vertice_{0}$, and finishes at some end vertex $\noCaminho_{w} = \vertice_{end} \in \ConjClusters_{end}$. 




Note that when each cluster has only one subgroup, the COPS reduces to the COP problem. On the other hand, when all subgroups are formed by a single vertex, the COPS simplifies to the SOP. Finally, if all clusters have one subgroup, and each subgroup has one vertex, then the COPS becomes equivalent to the classical OP.

\section{Methodology}
\label{section_methodology}

In order to solve the COPS we propose two approaches inspired by the ones in \cite{Angelelli2014Clustered}, an Integer Linear Program (ILP) formulation and a metaheuristic solution based on Tabu Search. The ILP can optimally solve small instances, and the Tabu Search is able to work with larger instances.

\subsection{ILP formulation}


This section presents a solution to the COPS based on an integer linear programming (ILP) formulation, as follows:
\begin{equation}
\label{eq:1}
    \max \sum_{\subgrupo_{i}~\in~\ConjSubgrupos} r_{i}z_{i} ~, 
\end{equation}
%
%
subject to the following constraints:
\begin{subequations}
\begin{equation}
    \label{eq:2}
    y_{1} = 1
\end{equation}
\begin{equation}
   \label{eq:3}
   \sum_{e \in \delta(j)} x_{e} = 2y_{j} \qquad \forall \vertice_{j} \in \ConjVertices_{client}
\end{equation}
\begin{equation}
    \label{eq:4}
    \sum_{e \in \delta(\cluster_{1})} x_{e} \leq 2
\end{equation}
\begin{equation}
     \label{eq:5}
     \sum_{e \in E} t_{e} x_{e} \leq \budget
\end{equation}

\begin{equation}
     \label{eq:6}
    \sum_{e \in \delta (U):} x_{e} \geq (\sum_{\vertice_{j} \in U } y_{j}) / |U|, \quad
    \forall ~U \subseteq \ConjVertices_{client}
\end{equation}
\begin{equation}
     \label{eq:7}
    z_{i} \leq y_{j}, \qquad \forall ~\subgrupo_{i} \in \ConjSubgrupos, \quad \forall \vertice_{j} \in \subgrupo_{i}
\end{equation}
\begin{equation}
     \label{eq:new_1}
    \sum_{\subgrupo_{i} \in \cluster_{g}} z_{i} = \visitedCluster_{g} \qquad \forall ~\cluster_{g} \in \ConjClusters
\end{equation}
\begin{equation}
    \label{eq:new_2}
    \sum_{\vertice_{i} \in \ConjVertices} y_{i} \leq \sum_{\subgrupo_{i} \in \ConjSubgrupos}Z_{i} |\subgrupo_{i}|
\end{equation}
\begin{equation}
    \label{eq:new_3}
    \visitedCluster_{g} \in \{0, 1\}, \quad \forall ~\cluster_{g} \in \ConjClusters
\end{equation}
\begin{equation}
    \label{eq:11}
    z_{i} \in \{0, 1\}, \quad \forall ~\subgrupo_{i} \in \ConjSubgrupos
\end{equation}
\begin{equation}
    \label{eq:12}
    x_{e} \in \{0, 1\}, \quad \forall ~e \in E
\end{equation}
\begin{equation}
    \label{eq:13}
    y_{j} \in \{0, 1\}, \quad \forall ~\vertice_{j} \in \ConjVertices
\end{equation}
\end{subequations}














In the case of non-circular route, add:
\begin{subequations}
\begin{equation}
    \label{eq:non-circular1}
     \sum_{\vertice_{j} \in \cluster_{end}} y_{j} = 1
\end{equation}
\begin{equation}
    \label{eq:non-circular2}
    \sum_{e \in \delta(\cluster_{end})} x_{e} = 1
\end{equation}
\end{subequations}





Where:

\begin{itemize}
  \item $\delta (U):$ group of edges with one extreme in $U$ and another in $\ConjVertices \setminus \{U\}$. 
  
  
  \item $x_{e}:$ binary variable with value 1 if the edge $e \in E$ is visited, and 0 otherwise.

  \item $y_{j}:$ binary variable with value 1 if the vertex $\vertice_{j} \in \ConjVertices$ is visited, and 0 otherwise.

  \item $z_{i}:$ binary variable with value 1 if all vertices of subgroup $\subgrupo_{i} \in \ConjSubgrupos$ are visited, and 0 otherwise.
    
  \item $\visitedCluster_{g}$: binary variable with value 1 if any subgroup of cluster $\cluster_{g} \in \ConjClusters$ is visited, and 0 otherwise.
\end{itemize}


This formulation is an adaptation of the one proposed for the COP in \cite{Angelelli2014Clustered}. The differences are related to the inclusion of the cluster class $\cluster_{g} \in \ConjClusters$; the binary variable described in \eqref{eq:new_3} and the new constraints \eqref{eq:new_1} and \eqref{eq:new_2}. 

Constraint \eqref{eq:new_1} causes the choice of at most one subgroup $\subgrupo_{i}$ for each cluster $\cluster_{g}$. If that cluster is being served, $\visitedCluster_{g} = 1$, then only one subgroup $\subgrupo_{i}$ can be selected. Note that, with just constraint \eqref{eq:new_1} the mathematical formulation already meets the requirements of the problem, and  constraint \eqref{eq:new_2} comes to minimize the energy expenditure of the path found, since it prevents additional vertices which do not reward the solution to be unnecessarily added to the path. In this case, the number of visited vertices must be less than or equal to the sum of the sizes of the visited subgroups. However, it can be smaller due to the case in which the same vertex belongs to more than one subgroup.

Constraints \eqref{eq:non-circular1} and \eqref{eq:non-circular2}  were also added to the formulation. These constraints guarantee that in the case where the final cluster differs from the initial cluster that only one vertex of the final cluster will be served \eqref{eq:non-circular1}, and the sum of the edges that have one endpoint at some vertex that belongs to the final cluster is equal to 1 \eqref{eq:non-circular2}.

Furthermore, there are some differences in constraints \eqref{eq:3}, \eqref{eq:4} and \eqref{eq:6}. In \cite{Angelelli2014Clustered} there is only one constraint that assures that for all visited vertices, there are two visited edges with an endpoint at that vertex. But in this work, it is assumed that the final vertex can differ from the initial vertex. In this case, constraint \eqref{eq:3} comes to guarantee that two edges are visited for each client vertex, and two or less in case the vertex is the initial, see constraint \eqref{eq:4}. It is noteworthy that by constraint \eqref{eq:2} the initial vertex will be in the solution. And constraint \eqref{eq:6} are the subtour elimination.


The others constraints are equal to the COP formulation \cite{Angelelli2014Clustered}. The constraint \eqref{eq:5} ensures that the path cost does not exceed the budget. Constraint \eqref{eq:7} imposes that all vertices belonging to a visited subgroup must be served. At last, \eqref{eq:11}, \eqref{eq:12} and \eqref{eq:13} are definitions of the binary variables.

Finally, the objective function \eqref{eq:1} maximizes the reward obtained by visiting the subgroups.

\subsection{Tabu Search metaheuristic}

The designed metaheuristic for the COPS is based on the Tabu Search \cite{glover1986future} method (Algorithm~\ref{alg:cap}). The algorithm starts choosing an initial solution, and at each iteration, the search moves to the best solution in the neighborhood, not accepting moves that lead to solutions already visited for a certain number of iterations; these known moves remain stored in a \emph{tabu list}. The Tabu Search was used to solve the COP problem in \cite{Angelelli2014Clustered}, and our approach is inspired by that.

\begin{algorithm}[htpb]
\caption{Tabu Search for the COPS}\label{alg:cap}
    \begin{algorithmic}[1]
        \STATE Compute an initial solution $\Caminho0$
        \STATE $\Caminho \gets \Caminho0$
        \STATE $\Caminho^* \gets \Caminho0$
        \STATE Initialize the long-term memory
        \WHILE{iterations without improvement $\leq \beta$}
            \STATE Update the tabu list
            \STATE Generate a plausible neighbor $\Caminho'$ from $\Caminho$
            \STATE $\Caminho \gets \Caminho'$
            \IF{$\Caminho'$ is better then $\Caminho^*$}
                \STATE $\Caminho^* \gets \Caminho$
            \ENDIF
            \STATE Update the long-term memory
            \IF{non-circular path not improved for $T$ steps}
                \STATE Change the final subgroup 
            \ENDIF
        \ENDWHILE
        \RETURN $\Caminho^*$
    \end{algorithmic}
\end{algorithm}

\subsubsection{Initial solution \Caminho0}

For the initial solution $\Caminho0$, we randomly choose one cluster at a time, and for each cluster, we select the subgroup with the highest profitability level. The profitability level is defined here as the profit achieved by visiting the subgroup divided by the number of vertices of this subgroup.
Then, this subgroup is inserted into $\Caminho0$, and a new tour is generated using the 2-opt strategy \cite{croes1958method}. The procedure stops when all clusters have been considered or the tour surpasses the budget for $\Lambda$ consecutive iterations.


\subsubsection{Long-term memory }

This is a memory, nominated here by $\eta_{i}$, that stores the number of iterations that a subgroup $\subgrupo_{i}$ has remained in or out to the current solution $\Caminho$. It will be set to one when $\subgrupo_{i}$ is inserted in $\Caminho$, and set to zero every time it is removed from $\Caminho$, thenceforth, it will be respectively increasing or decreasing in each iteration.

\subsubsection{Tabu list}


A subgroup will be a \emph{tabu} (forbidden) to insert or remove if, respectively, $\eta_{i} > -\alpha$ and $\eta_{i} < \alpha$, where $\alpha$ is the tabu constant.


\subsubsection{Generate a plausible neighbor $\Caminho'$}

Each iteration will generate a group of neighbors of the current solution $\Caminho$ in a predefined sequence. In order to avoid solving many TSP instances and then choose the best move, the first plausible solution found should be considered as the neighbor $\Caminho'$ for each iteration. Next, this neighbor should be compared with the final solution $\Caminho^*$. The final solution will be exchanged for this neighbor if it is more profitable as the final solution or has the same reward but with a lower traversal cost. Note that it will choose more efficient paths when the reward is the same but the cost is lower.

The rules to produce the neighbors follow two lines: (i) an insertion of some feasible subgroup in the current solution; or (ii) a removal of a subgroup from the solution.
The insertion is made by solving the classic local search strategy 2-opt \cite{croes1958method} for all chosen vertex.
The removal of a subgroup is further time optimized by just removing their vertices from the solution and join its predecessor to the successor. However, it is necessary to take care not to remove a vertex which is shared by any other subgroup in the current solution $\Caminho$.
Neighbors are generated in the following sequence:

\paragraph{Non-Tabu Insertion}

Choose randomly from $\ConjSubgrupos_{client}$ a subgroup that are not in the tabu list and that does not belong to a cluster that are in the current solution $\Caminho$, and insert it into the neighbor $\Caminho'$. 


\paragraph{Old Removal}

Remove from $\Caminho$ a randomly chosen subgroup from $\ConjSubgrupos_{client}$ for which $\eta_{i} > \beta$, where $\beta$ is a constant. 


\paragraph{Tabu Insertion}

Insert a subgroup from tabu insert list that does not belong to some cluster served in the current solution. The chosen subgroup is the one with the highest aspiration level. Similar to \cite{Angelelli2014Clustered}, the aspiration level is the highest reward value obtained on any solution that contained this subgroup.


\paragraph{Random Insertion}

Randomly chose and insert any subgroup from $\ConjSubgrupos_{client}$ that does not belong to some cluster served in the current solution.

\paragraph{Non-Tabu Removal}

Remove from $\Caminho$ a randomly chosen subgroup from $\ConjSubgrupos_{client}$ that is non-tabu to remove.

\paragraph{Random Removal}

Chose randomly a client subgroup from the current solution and remove it.

\subsubsection{Non-circular paths}

If the starting point differs from the ending point of the path, then there will be an end cluster, and all end subgroups must contain only one vertex, and the solver will choose which subgroup will be the end depot.

In this case, the endpoint will be changed after $T$ iterations without improvement. For this, we choose the subgroup with the minimum long-term memory, that is, the subgroup that has been outside the solution for the longest time. $T$ is defined as $\beta$ divided by the number of final subgroups. The idea is that the neighborhoods of all endpoints will be tested before the end of the run.

\subsubsection{Stop condition}

The algorithm stops after $\beta$ iterations without increasing the reward of the final solution, or decreasing the distance traveled, keeping the same reward. The value of $\beta$ obtained after preliminary tests was 300.

\section{Experiments} \label{section_Experiments}



The algorithms\footnote{Code available in: \url{https://github.com/verlab/COPS}} were implemented in Python~3.7, and the ILP uses Gurobi\footnote{\url{https://www.gurobi.com/}} 10.0.0 Academic License as the solver. All tests have been made on a machine with Intel Xeon E5-2630 v3 2.4GHz, with 128GB RAM.

The conducted experiments have three main objectives:
\begin{itemize}
    \item To verify the applicability of the COPS approach to solve both COP and SOP variants;
    \item To explore new possibilities of application enabled by the specific COPS formulation;
    \item To evaluate the performance of the proposed algorithms.
\end{itemize}
%


\subsection{Classic COP and SOP instances} 


Initially, we aim to assess the COPS approach's efficacy in addressing classic COP and SOP variants. To accomplish this, we had to adapt the original benchmark instances to allow for comparing the solutions.



\begin{figure}[htpb]
    \centering
    \subfigure[SOP format.]{
        \includegraphics[trim={1.75cm 0 1.75cm 1.25cm},clip,height=5cm]{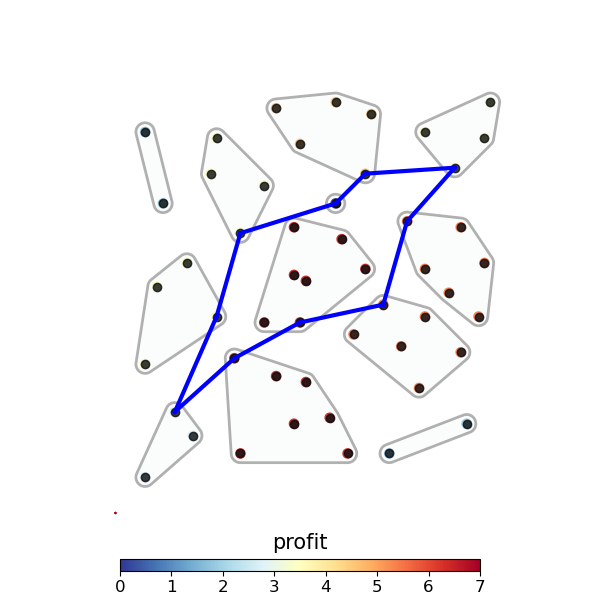}
        \label{fig:sop_iel}
    }
    \hspace{-.5cm}
    \subfigure[COP format.]{
        \includegraphics[trim={1.75cm 0 1.75cm 1.25cm},clip,height=5cm]{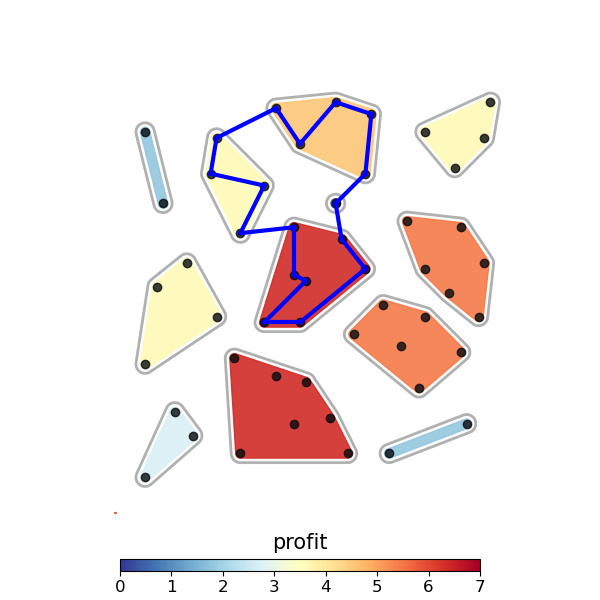}
        \label{fig:cop_iel}
    }
    \caption{Adapted 11iel51 instance with $\budget = 139$.}
    \label{fig:SOP_e_COP}
\end{figure}

We modified the SOP instances in~\cite{Archetti2018Set} by making each vertex a subgroup with the original reward and defining like a cluster each group called \emph{set} in~\cite{Archetti2018Set}. Note that our method allows the creation of SOP instances with vertices with different rewards. In Table~\ref{tab:sop}, we compare the total reward $\rewardEarned$ and runtime $\runtime$ (seconds) for the MASOP~\cite{Archetti2018Set}, the VNS-SOP~\cite{Penicka2019Variable}, and our approach, Fig.~\ref{fig:sop_iel} shows an example where the route generated by our Tabu Search achieves a higher reward than both MASOP and VNS-SOP. 

\begin{table}[htpb]
\centering
\caption{Results for experiment comparing with SOP solutions.}
\label{tab:sop}
\resizebox{\linewidth}{!}{
    \begin{tabular}{|c|c|c|c|c|c|c|c|c|}
        \hline
        \multirow{2}{*}{\textbf{Instance}} & \multirow{2}{*}{\textbf{Size}} & \multirow{2}{*}{\textbf{$\budget$}} & \multicolumn{2}{c|}{\textbf{MASOP}~\cite{Archetti2018Set}} & \multicolumn{2}{c|}{\textbf{VNS-SOP}~\cite{Penicka2019Variable}} & \multicolumn{2}{c|}{\textbf{COPS-TABU}} \\ \cline{4-9} 
        
         &  & & \textbf{$\rewardEarned$} & \textbf{$\runtime$} & \textbf{$\rewardEarned$} & \textbf{$\runtime$} &  \textbf{$\rewardEarned$} & \textbf{$\runtime$} \\ \hline
         
        11berlin52 & 52 & 1616 & 37 & 1.75 & 37 & 0.11 & 37 & 1.1\\ \hline
        11berlin52 & 52 & 2424 & 43 & 2.4 & 43 & 0.16 & 43 & 1.3\\ \hline
        11berlin52 & 52 & 3232 & 47 & 4.63 & 47 & 0.19 & 47 & 3.7\\ \hline

        11eil51 & 51 & 69 & 24 & 1.85 & 24 & 0.09 & 24 & 0.4\\ \hline
        11eil51 & 51 & 104 & 39 & 5.13 & 39 & 0.14 & 39 & 1.4\\ \hline
        11eil51 & 51 & 139 & 43 & 2.30 & 43 & 0.18 & \textbf{46} & 3.\\ \hline
        
        26bier127 & 127 & 72418 & 125 & 16 & 125 & 1.8 & 125 & 70\\ \hline
         
    \end{tabular}
    }
\end{table}

Fig.~\ref{fig:cop_iel} shows the same instance but now formulated as a COP instance. Note that, here, all cluster points belong to the same subgroup; in other words, we have only one subgroup for each cluster. In Table~\ref{tab:cop}, we consider traditional instances for the COP, presented in~\cite{Angelelli2014Clustered}, with their results, denoted COP-TABU, and compare with our formulation.

\begin{table}[htpb]
\centering
\caption{Results for experiment comparing with COP solutions.}
\label{tab:cop}
\resizebox{\linewidth}{!}{
    \begin{tabular}{|c|c|c|c|c|c|c|}
        \hline
        \multirow{2}{*}{\textbf{Instance}} & \multirow{2}{*}{\textbf{Size}} & \multirow{2}{*}{\textbf{$\budget$}} & \multicolumn{2}{c|}{\textbf{COP-TABU}~\cite{Angelelli2014Clustered}} & \multicolumn{2}{c|}{\textbf{COPS-TABU}} \\ \cline{4-7} 
        
         & & & \textbf{$\rewardEarned$} & \textbf{$\runtime$} &  \textbf{$\rewardEarned$} & \textbf{$\runtime$} \\ \hline
         
        eil51s25g1q2 & 51 & 229 & 32 & 94.64 & \textbf{36} & 89.84\\ \hline
        berlin52s10g1q2 & 52 & 3785 & 21 & 2.69 & 21 & 35.92\\ \hline
        st70s25g1q2 & 70 & 355 & 34 & 112.18 & 34 & 84.14\\ \hline
        eil76s25g1q2 & 76 & 292 & 50 & 113.00 & 50 & 281.6\\ \hline
        pr76s25g1q2 & 76 & 54096 & \textbf{65} & 125.90 & 60 & 357.82\\ \hline
        kroC100s25g1q2 & 100 & 10401 & 30 & 87.36 & 30 & 104.3\\ \hline
        kroE100s25g1q2 & 100 & 11061 & \textbf{36} & 118.07 & 30 & 67.62\\ \hline
        rd100s25g1q2 & 100 & 3979 & 42 & 193.65 & 42 & 227.00\\ \hline
    \end{tabular}
    }
\end{table}





We checked through the paired t-test with 95\% confidence that the COPS-TABU manages to find similar results, in terms of reward, to the state-of-the-art considering both COP and SOP.
Regarding computational time, the comparison can only be partially analyzed due to the different settings used to perform the experiments. However, it can be seen that our approach obtains worse results when compared to VNS-SOP, especially for larger instances, \eg 26bier127 of Table~\ref{tab:sop}. In comparison with COP-TABU, our algorithm obtains similar results. Note that this work does not intend to have the most optimized method in terms of execution time; however, we provide a performance analysis in Sec.~\ref{secao:performance_analysis}.

\subsection{COPS-specific scenarios}

\begin{figure}[htpb]
    \centering
    \subfigure[$\budget = 25$]{
       \includegraphics[height=2.55cm]{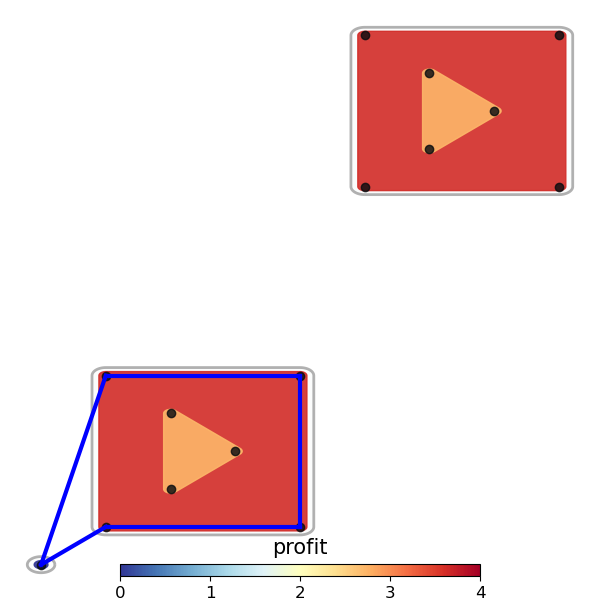}
    }
    \subfigure[$\budget = 35$]{
        \includegraphics[height=2.55cm]{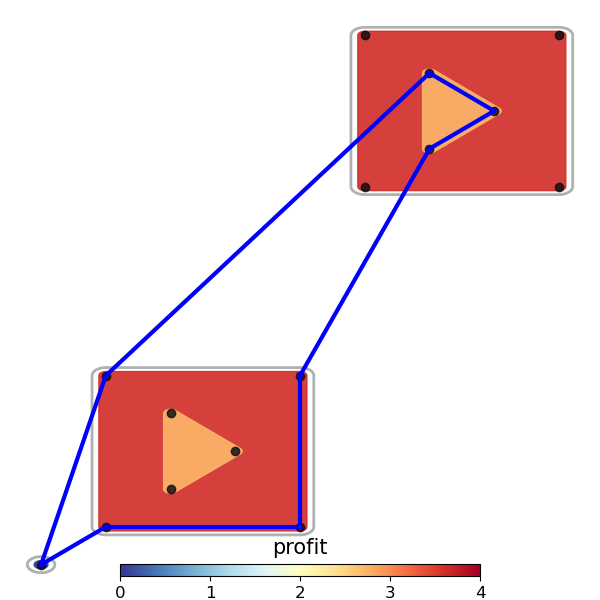}
    }
    \subfigure[$\budget = 40$]{
        \includegraphics[height=2.55cm]{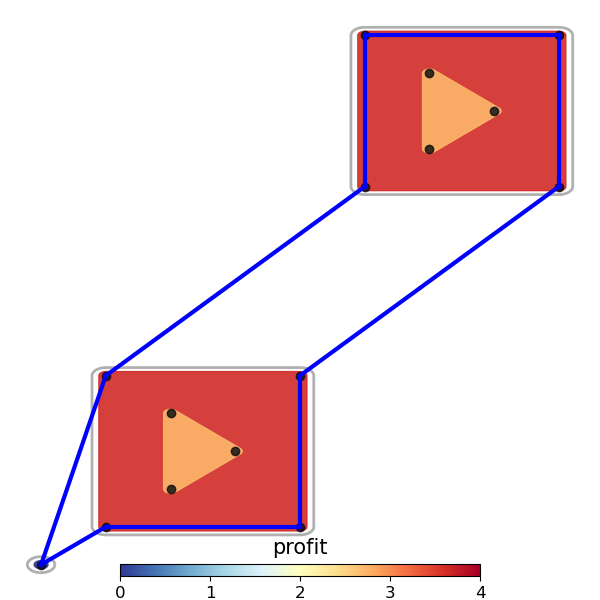}
    }

    
    \caption{Results for COPS-TABU varying robot budget.}
    \label{fig:varying_robot_budget}
\end{figure}

This section explores a simple example where we vary the travel budget to illustrate the COPS functionalities. This example considers two clusters, each with two subgroups (Fig.~\ref{fig:varying_robot_budget}). Visiting the largest subgroups requires a longer path, but they give a greater reward. We aim to show that we can choose subgroups within each cluster to achieve larger earnings while respecting the budget. Sometimes, choosing the less profitable subgroup to visit more clusters is better.



Table~\ref{tab:varying_robot_budget} contains the results for both methods and a larger range of travel budgets. Note that the path length ($\length$) obtained by the ILP and the metaheuristic may differ, even achieving the same reward. This happens as the OP's objective is to maximize reward, but not necessarily decrease the budget utilized. We also highlight that for certain budgets (\eg, $\budget=10$) it might be unfeasible to find a solution.




\begin{table}[htpb]
\centering
\caption{Results for experiment varying robot budget.}
\label{tab:varying_robot_budget}
    \begin{tabular}{|c|c|c|c|c|c|c|}
        \hline
        \multirow{2}{*}{\textbf{$\budget$}} & \multicolumn{3}{c|}{\textbf{ILP}} & \multicolumn{3}{c|}{\textbf{COPS-TABU}} \\ 
        
        \cline{2-7}

         & \textbf{$\rewardEarned$} & \textbf{$\length$} & \textbf{$\runtime$} & \textbf{$\rewardEarned$} & \textbf{$\length$} & \textbf{$\runtime$} \\ \hline
        
        40 & 8 & 39.6 & 2.7 & 8 & 37.3 & 0.3 \\ \hline
        35 & 7 & 32.4 & 10.5 & 7 & 32.1 & 0.5\\ \hline
        30 & 6 & 29.9 & 18.8 & 6 & 29.9 & 0.5 \\ \hline
        25 & 4 & 16.5 & 7.3 & 4 & 16.5 & 0.6 \\ \hline
        20 & 4 & 16.5 & 6.4 & 4 & 16.5 & 0.5 \\ \hline
        15 & 3 & 10.1 & 10.8 & 3 & 10.4 & 0.3 \\ \hline
        10 & 0 & 0 & 4.5 & 0 & 0 & 0.2 \\ \hline
        
    \end{tabular}
\end{table}



\begin{figure}[htpb]
    \centering
    \subfigure[]{
        \includegraphics[height=2.5cm]{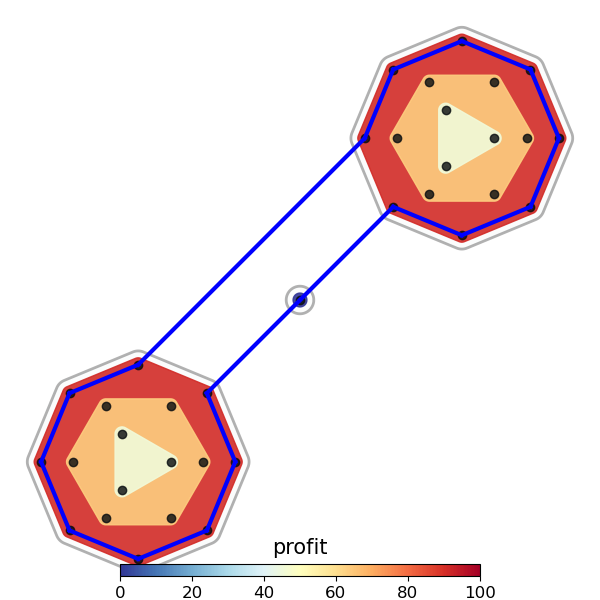}
        \label{fig:varying_number_of_groups_a}
    }
    \subfigure[]{
        \includegraphics[height=2.5cm]{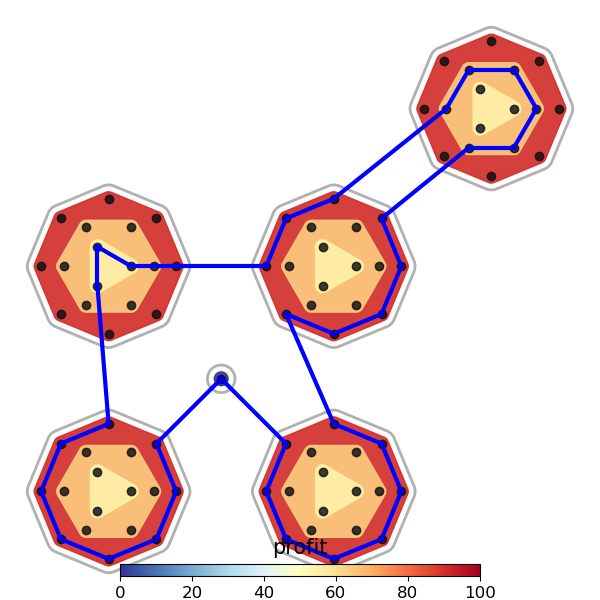}
        \label{fig:varying_number_of_groups_b}
    }
    \subfigure[]{
        \includegraphics[height=2.5cm]{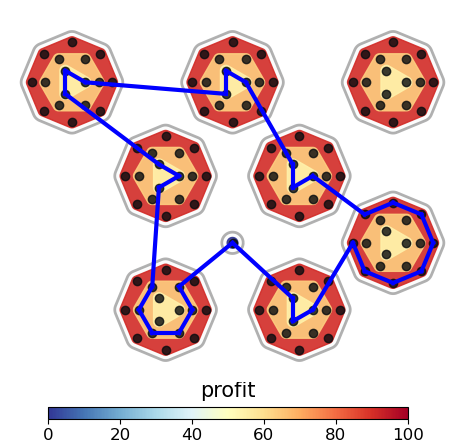}
        \label{fig:varying_number_of_groups_d}
    }
    \caption{Results for COPS-TABU varying number of clusters.}
    \label{fig:varying_number_of_groups}
\end{figure}

Next, we fix the budget at $\budget=100$, and progressively increase the number of clusters, verifying which clusters/subgroups are visited and the reward earned. Each cluster has three subgroups with different associated rewards, and Fig.~\ref{fig:varying_number_of_groups} illustrates some of the cases.
We detail the results in Table~\ref{tab:varying_number_of_groups}; as expected, the collected reward increases with the addition of more clusters until the travel budget limits the expansion of the route. Furthermore, Figs.~\ref{fig:varying_number_of_groups_b} and \ref{fig:varying_number_of_groups_d} show that our approach can efficiently select all three types of subgroups.
The ILP is not used in this case due to the model's size surpassing the RAM limits (details in Section~\ref{secao:performance_analysis}).

\begin{table}[htpb]
\centering
\caption{Results for experiment varying number of clusters.}
\label{tab:varying_number_of_groups}
    \begin{tabular}{|c|c|c|c|c|}
        \hline
        \multirow{2}{*}{\textbf{Clusters}} & \multirow{2}{*}{\textbf{$\budget$}} & \multicolumn{3}{c|}{\textbf{COPS-TABU}} \\ \cline{3-5}

         & & \textbf{$\rewardEarned$} & \textbf{$\length$} & \textbf{$\runtime$} \\ \hline
        
        1 & 100 & 100 & 25.5 & 0.24 \\ \hline
        2 & 100 & 200 & 50.18 & 2.27 \\ \hline
        3 & 100 & 300 & 67.4 & 9.07 \\ \hline
        4 & 100 & 400 & 89.2 & 15.2 \\ \hline
        5 & 100 & 430 & 69.4 & 81.52 \\ \hline
        6 & 100 & 450 & 97.14 & 79.77 \\ \hline
        7 & 100 & 470 & 98.27 & 146.22 \\ \hline
        8 & 100 & 470 & 98.1 & 158.61 \\ \hline

    \end{tabular}
\end{table}



\subsection{Performance analysis}
\label{secao:performance_analysis}

To comprehensively understand the methods' behavior in scaling scenarios, we conducted a statistical analysis to assess how the total number of vertices, subgroups, and clusters impact the performance. Throughout these experiments, the budget ($T_{max}$) consistently allowed for ample coverage of the entire environment.



\begin{figure}[htpb]
    \centering
    \includegraphics[width=.9\linewidth]{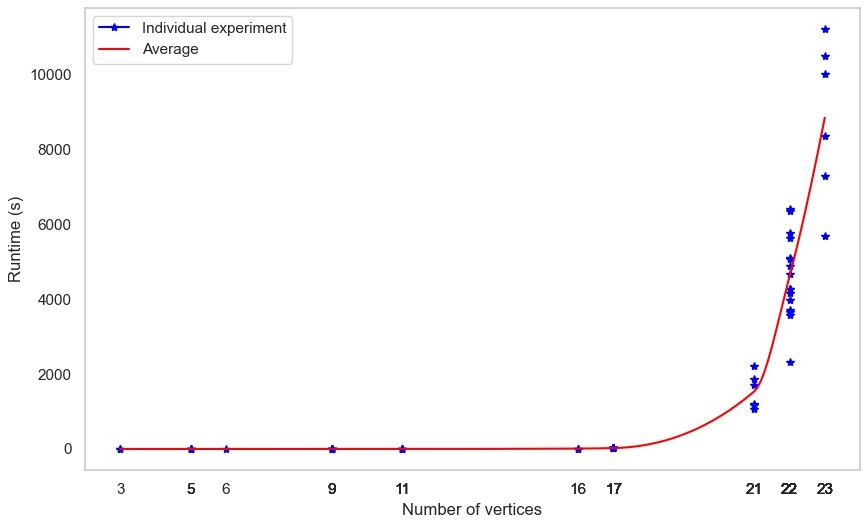}
    \caption{ILP performance analysis with increasing number of vertices across different configurations of clusters and subgroups.}
    \label{fig:analise_ilp}
\end{figure}

Initially, considering the ILP formulation, we systematically increased the number of vertices across different configurations of clusters and subgroups. The results revealed a significant correlation between computational time and the number of vertices, regardless of cluster or subgroup quantities (Fig.~\ref{fig:analise_ilp}). We replicated the experiment for COPS-TABU, and results are presented in Table~\ref{tab:ilp_vs_cops-tabu}. This comparison indicates that ILP and COPS-TABU perform similarly for lower vertex counts, while COPS-TABU exhibits superior performance as vertex counts increase.
The experiment reached a limit of 23 vertices due to resource constraints, specifically, RAM consumption during the experiments exceeded the available memory capacity. Although the evaluation was limited, this restriction did not compromise the integrity of our discoveries, but highlights the limited usage of the ILP.

\begin{table}[htpb]
\centering
\caption{Results for experiment comparing the ILP with COPS-TABU.}
\label{tab:ilp_vs_cops-tabu}
    \begin{tabular}{|c|c|c|c|c|}
        \hline
        \multirow{2}{*}{\textbf{Vertex}} & \multicolumn{2}{c|}{\textbf{ILP}} & \multicolumn{2}{c|}{\textbf{COPS-TABU}} \\ \cline{2-5} 
        
          & \textbf{$\runtime$} & 
         RAM (GB) &
         \textbf{$\runtime$} & RAM (GB) \\ \hline

3 & 0.03 & 0.50 & 0.02 & 0.1 \\ \hline
5 & 0.03 & 0.49 & 0.09 & 0.11 \\ \hline
6 & 0.06 & 0.50 & 0.7 & 0.13 \\ \hline
9 & 0.07 & 0.54 & 0.4 & 0.11 \\ \hline
11 & 0.25 & 0.54 & 0.7 & 0.12 \\ \hline
16 & 11.17 & 0.64 & 0.4 & 0.12 \\ \hline
17 & 24.5 & 0.84 & 2.7 & 0.11 \\ \hline
21 & 1540.72 & 20.56 & 7.46 & 0.13 \\ \hline
22 & 4618.92 & 45.6 & 7.24 & 0.11 \\ \hline
23 & 8854.56 & 96.44 & 8.57 & 0.12 \\ \hline
         
    \end{tabular}
\end{table}


To bolster our findings with a more rigorous approach, we employed a $2^k$ experimental design with 30 replications. Here, `$k$' denotes the number of factors or variables under examination -- in our case, $k = 3$. Each factor is tested at two levels (high and low), encompassing all possible combinations of these levels. Specifically, our experiment involved testing the number of clusters at 2 and 4, subgroups at 4 and 8, and vertices at 8 and 16. The primary objective of this factorial design is to identify which factors and their interactions notably impact runtime.


The analysis showed that among the factors explored, the number of vertices dominates the observed variations in execution time, accounting for approximately 89.44\% of the variance. 
In contrast, the number of clusters and subgroups exhibited minimal effects at 0.02\% and 0.001\%, respectively, while their interactions collectively contributed 0.039\% to the overall variance.
This indicates a limited impact compared to the substantial influence exerted by the vertex count. Notably, approximately 10.5\% of the observed variations relates to experimental error, emphasizing the importance of controlling additional variables in future investigations.

Finally, we investigated the impact of the total number of vertices, clusters, and subgroups on COPS-TABU. This experiment unfolds across four stages: in stages 1 and 2, we maintain the cluster count at 10 while progressively increasing subgroups from 30 to 60; in stages 3 and 4, we elevate the cluster count to 15 while reevaluating the previously tested subgroup quantities. Our assessment involves analyzing COPS-TABU's runtime through a sequence of runs, commencing at 60 vertices and incrementing by 10 until approaching a runtime of 1500 seconds (with $\budget=1000$, we effectively encompass all feasible path scenarios).

\begin{figure}[htpb]
    \centering
    \includegraphics[width=.85\linewidth]{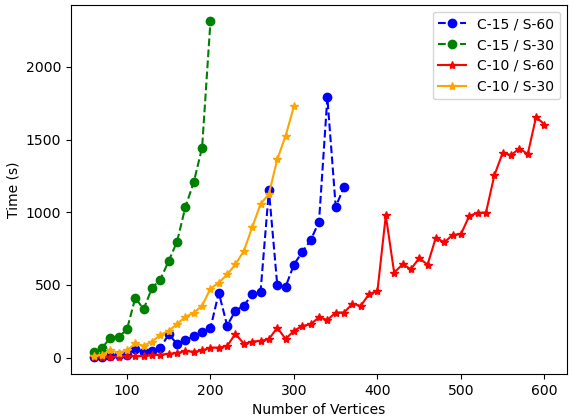}
    \caption{Analysis of the runtime behavior of the COPS-TABU algorithm while varying the number of vertices, subgroups, and clusters.}
    \label{fig:tab:projeto_2k}
\end{figure}

Regarding the influence of the number of clusters, Fig.~\ref{fig:tab:projeto_2k} shows a direct proportionality with execution time. This correlation emerges because Tabu Search needs to explore a larger neighborhood to converge to a local maximum. The comparison among scenarios with the same number of subgroups but differing cluster counts demonstrates that a higher number of clusters corresponds to longer runtimes.
On the other hand, the runtime shows an inverse correlation with the number of subgroups, evident from the distinction between the blue and green lines, as well as the red and orange lines.
The rationale behind this trend is the fact that fewer subgroups imply a higher concentration of vertices within each subgroup. Consequently, the 2-opt strategy must evaluate solutions encompassing more vertices, leading to extended response times.


Expanding our investigation, we conducted a $2^k$ factorial experiment to delve deeper into the behavior of the COPS-TABU algorithm. This experiment involves cluster levels set at 10 and 15, subgroup levels at 30 and 60, and vertex levels at 100 and 200. The goal was to evaluate how these factors, both individually and in interaction, influence the runtime.


Notably, the number of vertices emerges as the most significant contributor, accounting for approximately 22.35\% of the overall variance. Subsequently, both the number of subgroups at 21.37\% and the number of clusters at 13.18\% also exhibit considerable individual influences on runtime.
Moreover, the collective interactions between these factors also contribute notably, underscoring their combined impact on runtime. 
For instance, the interaction between clusters and subgroups amounts to 9.57\%, clusters and vertices show 9.52\%, and subgroups and vertices demonstrate 15.53\%. Simultaneously, the interaction involving all three factors contributes 7.34\% to the overall variance.


These findings from the $2^k$ factorial experiment provide valuable insights into how variations in the number of clusters, subgroups, and vertices, as well as their interactions, significantly influence the runtime performance of COPS-TABU. The relatively low percentage of experimental error at 1.15\% emphasizes the robustness and reliability of these observed effects in our study.

\section{Conclusion and Future Work}


This paper introduces a novel generalization of the \ac{OP} called \ace{COPS}.
In the COPS, vertices are arranged into subgroups, which are combined into clusters. This new variant manages to address the classic OP generalization, as well as the COP and SOP variations.
In addition, this new formulation can be used to model scenarios that have never been addressed in the literature, such as heterogeneous subgroups within a cluster. Furthermore, other cases are also allowed and have not been presented here due to space limitations, such as endpoints with different rewards, shared vertices between subgroups, and shared subgroups between clusters, among others. 

We proposed two approaches to tackle this new problem, an exact ILP method and a Tabu Search-based heuristic. Experimental findings suggest that the ILP approach can deliver solutions with optimality guarantees with a high computational time, whereas the Tabu Search can provide solutions of similar quality within a more practical runtime.

In future work, we intend to evaluate the use of our formulation to solve other variants, such as the Orienteering Problem with Neighborhoods (OPN) and the Dubins Orienteering Problem (DOP). We also plan to improve the performance of our exact method, for example, by introducing a branch-and-bound strategy. Finally, we consider extending the formulation to multi-robot assembles.





\bibliographystyle{IEEEtran}
\bibliography{bibliography}

\end{document}